# Speedy Model Selection (SMS) for Copula Models

**Yaniv Tenzer and Gal Elidan**
Department of Statistics, The Hebrew University

## Abstract

We tackle the challenge of efficiently learning the structure of expressive multivariate real-valued densities of copula graphical models. We start by theoretically substantiating the conjecture that for many copula families the magnitude of Spearman's rank correlation coefficient is monotonic in the expected contribution of an edge in network, namely the negative copula entropy. We then build on this theory and suggest a novel Bayesian approach that makes use of a prior over values of Spearman's rho for learning copula-based models that involve a mix of copula families. We demonstrate the generalization effectiveness of our highly efficient approach on sizable and varied real-life datasets.

## 1 Introduction

Learning expressive real-valued multivariate distributions is of central interest in numerous fields ranging from computational biology to economics to climatology. When the joint distribution of interest is far from multivariate normal, this modeling task can be a great challenge. In statistics, copulas [Joe, 1997, Nelsen, 2007] are the central tool for capturing flexible multivariate real-valued distributions by separating the choice of the univariate marginals and the copula function that links them. Copulas are typically only effective in low dimensions, and much of the research in the field in fact focuses on the bivariate case. Accordingly, in the last decade, various high-dimensional constructions that build on a collection of copulas have been suggested, most notably those based on the vine construction [Bedford and Cooke, 2002, Kurowicka and Cooke, 2002]. These models have proved to be quite effective for crossing the *few variable* barrier. However, in the context of many tens of variables to hundreds and thousands of variables, applications have been few and involve costly and time-consuming expert elicitation [Hanea et al., 2010].

In machine learning, probabilistic graphical models, and in particular directed Bayesian networks (BNs) [Pearl, 1988], have become increasingly popular as a flexible and intuitive framework for modeling multivariate densities based on a qualitative graph structure $\mathcal{G}$ that encodes the independencies in the domain. Graphical models are geared toward the high-dimensional case and numerous algorithms for estimation, model selection and prediction using these models have been developed in recent decades [Koller and Friedman, 2009]. Unfortunately, due to computational considerations, real-valued high-dimensional modeling using this framework is often limited to a structured multivariate Gaussian model.

In recent years, several works suggested a fusion between copulas and graphical models [Kirshner, 2007, Elidan, 2010], with the goal of allowing for flexible real-valued modeling that is practical in the high-dimensional setting. The basic idea is that the joint density is defined via a collection of local copula functions that capture the direct dependence between a variable and its parents in the graph $\mathcal{G}$, as well as a set of univariate marginals that are shared across the entire model. As with standard graphical models, the super-exponential task of learning the structure of such models from data poses practical difficulties. Specifically, the computational bottleneck of structure learning is the assessment of the quality of candidate structures, which in turn requires costly estimation of maximum likelihood parameters. Indeed, even when using a simple greedy procedure to traverse the space of structures, or when limiting ourselves to tree structured models, structure learning can be computationally demanding for non-Gaussian models. Our goal in this work is to cope with this challenge.

Recently, Elidan [2012] suggested a highly efficient approach for learning the structure of copula-based

Bayesian networks. Briefly, the building block of structure learning is the *ranking* of the merit of an edge $X \rightarrow Y$ given $M$ training samples. Using $U \equiv F_X(x)$, $V \equiv F_Y(y)$ to denote the marginal distributions of these variables and assuming a copula-based model, they note that the benefit of the edge is asymptotically equal to the negative differential entropy

$$-H(c_\theta(U,V)) = \int c_\theta(u,v) \log c_\theta(u,v) du dv, \quad (1)$$

where $c(\cdot)$ denotes the (copula) density that corresponds to the joint distribution of $X$ and $Y$. They then suggest that $-H(c_\theta(U,V))$ is monotonic in the easy to compute Spearman's $\rho_s$ measure of correlation. This in turn facilitates highly efficient structure learning where simple Spearman's $\rho_s$ computations are used to rank candidate edge modifications. The monotonicity is proved for the Gaussian copula and algebraically simple Farlie-Gumbel-Morgenstern (FGM) copula family. Based on simulations, they further conjectured that the result holds for several additional copula families. Finally, they show that the method can be used to learn the structure of copula-based models that generalize well very efficiently. The method's main limitation, other than the gap in theory, is the fact that the same copula family is used to parameterize all edges in the model.

In this work we extend Elidan [2012] along two important axes. First, we provide a formal proof of the monotonicity conjecture given a sufficient condition that applies to a wide range of common copula families, a novel contribution to the theory of copulas on its own. Second, we tackle the challenge of performing structure learning while also allowing for a mixed combination of copulas, thereby significantly increasing the expressive power of the model. Briefly, our theoretical result suggests that the selection between copula families can be made based on expected likelihood *characteristic curves* that are computed *once*. A natural Bayesian prior is then used to "calibrate" the curves for several families. Finally, the posterior curves are used to select a copula family for each edge based *only* on Spearman's $\rho_s$ computations.

We use our speedy model selection (SMS) approach to learn copula-based tree structured models for several real-life datasets that are quite substantial in size in the context of structure learning with the number of variables ranging from 100 to close to 900. In all cases, we demonstrate impressive performance benefits relative to learning a model that is constrained to using a single copula family. Further, in many instances we show that our highly efficient approach is competitive with the computationally demanding golden standard where the best copula for each edge is computed via *costly* maximum likelihood estimation for each family.

## 2 Background

In this section we briefly provide the necessary background on copulas, Spearman's $\rho_s$ and stochastic orders of multivariate distributions.

### 2.1 Copula and Spearman's $\rho_s$

A copula function joins univariate marginals into a joint real-valued multivariate distribution. Formally,

**Definition 2.1:** Let $U_1, \ldots, U_n$ be random variables marginally uniformly distributed on $[0,1]$. A copula function $C : [0,1]^n \to [0,1]$ is a joint distribution

$$C_\theta(u_1, \ldots, u_n) = P(U_1 \leq u_1, \ldots, U_n \leq u_n),$$

where $\theta$ are the parameters of the copula function. ∎

Now consider an arbitrary set $\mathcal{X} = \{X_1, \ldots X_n\}$ of real-valued random variables (typically *not* marginally uniformly distributed). Sklar's seminal theorem [Sklar, 1959] states that for *any* joint distribution $F_\mathcal{X}(\mathbf{x})$, there exists a copula function $C$ such that

$$F_\mathcal{X}(\mathbf{x}) = C(F_1(x_1), \ldots, F_n(x_n)).$$

When the univariate marginals are continuous, $C$ is uniquely defined.

The constructive converse, which is of central interest from a modeling perspective, is also true. Since $U_i \equiv F_i$ is itself a random variable that is always uniformly distributed in $[0,1]$, *any* copula function taking *any* marginal distributions $\{F_i(x_i)\}$ as its arguments, defines a valid joint distribution with marginals $\{F_i(x_i)\}$. Thus, copulas are "distribution generating" functions that allow us to separate the choice of the univariate marginals and that of the dependence.

To derive the joint *density* $f(\mathbf{x}) = \frac{\partial^n F(x_1, \ldots, x_n)}{\partial x_1 \ldots \partial x_n}$ from the copula construction, assuming $F$ has n-order partial derivatives (true almost everywhere when $F$ is continuous), and using the chain rule, we have

$$f(\mathbf{x}) = \frac{\partial^n C(F_1(x_1), \ldots, F_n(x_n))}{\partial F_1(x_1) \ldots \partial F_n(x_n)} \prod_i f_i(x_i)$$
$$\equiv c(F_1(x_1), \ldots, F_n(x_n)) \prod_i f_i(x_i),$$

where $c(F_1(x_1), \ldots, F_n(x_n))$, is called the *copula density function*.

**Example 2.2:** The Gaussian copula is undoubtedly the most commonly used copula family and is defined as

$$C_\Sigma(\{U_i\}) = \Phi_\Sigma\left(\Phi^{-1}(U_1), \ldots, \Phi^{-1}(U_N)\right), \quad (2)$$

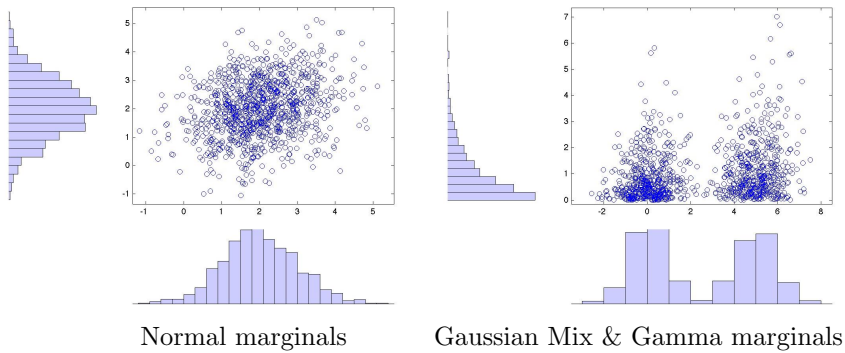

Figure 1: Samples from the bivariate Gaussian copula with correlation $\theta = 0.25$. (left) with unit variance Gaussian marginals; (right) with a mixture of Gaussian and Gamma marginals.

Normal marginals  Gaussian Mix & Gamma marginals

where $\Sigma$ is a correlation matrix, $\Phi$ is the standard normal distribution, and $\Phi_\Sigma$ is a zero mean normal distribution with correlation matrix $\Sigma$. Figure 1 exemplifies the flexibility that comes with this seemingly limited elliptical copula family.

Copula are intimately connected to many dependence concepts such as Spearman's $\rho_s$ measure of association

$$\rho_s(X_1, X_2) = \frac{cov(F_{X_1}, F_{X_2})}{STD(X_1)STD(X_2)},$$

which is simply Pearson's correlation applied to the cumulative distributions of $X_1$ and $X_2$. For the copula associated with the joint $F_{X_1,X_2}(x_1, x_2)$, we have

$$\rho_s(X_1, X_2) = \rho_s(C) \equiv 12 \int \int C(u,v) du dv - 3.$$

Thus, Spearman's $\rho_s$ is monotonic in the copula cumulative distribution function associated with the joint distribution of $X_1$ and $X_2$. See [Nelsen, 2007, Joe, 1997] for an in-depth exploration of the framework of copulas and its relationship to dependence measures.

### 2.2 Stochastic Orderings

The vast majority of copula families are parameterized by a dependence parameter $\theta$ that defines a stochastic ordering in the bivariate case. Below we define two such related orders that will be used in the sequel. In the rest of the paper, we use $\mathbf{X}$, $\mathbf{Y}$ to denote bivariate random vectors with distributions $F_\mathbf{X}(u,v)$ and $F_\mathbf{Y}(u,v)$, respectively.

**Definition 2.3:** $\mathbf{Y}$ is said to be more positive quadrant dependent than $\mathbf{X}$, denoted by $\mathbf{X} \leq_{PQD} \mathbf{Y}$, if $\forall (u,v) \in \mathbf{R}^2, F_\mathbf{X}(u,v) \leq F_\mathbf{Y}(u,v)$.

Thus, PQD ordering corresponds to a fast accumulation of density. To define the second ordering, we first need the notion of supermodularity. In what follows we use the following notation: $u \vee v \equiv min(u,v)$, $u \wedge v \equiv max(u,v)$.

**Definition 2.4:** A function $\Psi : \mathbf{R}^2 \to \mathbf{R}$ is said to be *supermodular* if

$$\forall (u,v) \in \mathbf{R}^2, \Psi(u \vee v) + \Psi(u \wedge v) \geq \Psi(u) + \Psi(v)$$

$\Psi$ is submodular when the inequality is reversed.

The supermodular ordering can now be defined:

**Definition 2.5:** $\mathbf{Y}$ is said to be greater than $\mathbf{X}$ in the supermodular order, denoted by $\mathbf{X} \leq_{sm} \mathbf{Y}$, if $\forall \Psi$ such that $\Psi$ is super modular: $E[\Psi(\mathbf{X})] \leq E[\Psi(\mathbf{Y})]$

This property is important in our context since, in the bivariate case, we have the following result due to Shaked and Shanthikumar [2007]:

**Theorem 2.6:** $\mathbf{X} \leq_{PQD} \mathbf{Y} \iff \mathbf{X} \leq_{sm} \mathbf{Y}$.

## 3 Monotonicity of the Copula Entropy in the Dependence Parameter

As discussed in the introduction, Elidan [2012] suggested that the magnitude of Spearman's $\rho_s$ is monotonic in the negative copula entropy which in turn asymptotically approximates the expected log-likelihood of a model, thereby giving rise to an efficient structure learning procedure. In this section we prove the conjecture for a wide range of copula families and discuss its relationship to real-valued majorization. In the next section we present a novel algorithmic approach called speedy model selection (SMS) that builds on this theory and allows us to efficiently perform structure learning while *at the same time* choose the local copula family.

### 3.1 TP2 Density Implies Entropy Ordering

We now present our central result, namely the identification of a widely applicable sufficient condition for the monotonicity of the copula entropy in the dependence parameter, and consequently in Spearman's $\rho_s$.

Recall that $\mathbf{X}$, $\mathbf{Y}$ are two bivariate random vectors. Throughout this section let $\mathbf{X} \sim C_{\theta_1}(u,v), \mathbf{Y} \sim$

$C_{\theta_2}(u,v)$ with $\theta_1 < \theta_2$, where $C_\theta(u,v)$ is an absolutely continuous bivariate copula family that is increasing in $<_{PQD}$ so that the cumulative distribution of $\mathbf{Y}$ is greater than that of $\mathbf{X}$ for all $(u,v)$. Note that essentially all copula families that are parameterized by a so called dependence parameter $\theta$ are PQD ordered.

Before stating the main result, using $u \vee v \equiv min(u,v)$, $u \wedge v \equiv max(u,v)$, we define the following notion:

**Definition 3.1:** A function $\Psi : \mathbf{R}^2 \to \mathbf{R}$ is TP2 (total positive of order 2) if the following holds:

$$\forall (u,v) \in \mathbf{R}^2 \quad \Psi(u \vee v) \cdot \Psi(u \wedge v) \geq \Psi(u) \cdot \Psi(v).$$

$\Psi$ is called RR2 (reversed regular of order 2) when the inequality is reversed. ∎

Note that the density for many copulas is a TP2 or RR2 function (for example, 8 of the twelve B1-B12 families defined in Joe [1997] are known to be TP2 and the property may hold for some of the others).

The following property of TP2 (RR2) functions, easily proved using logarithmic properties, will be needed:

**Observation 3.2:** Given a positive function $\Psi(u,v)$ which is TP2 (RR2), $\Phi(u,v) = log(\Psi(u,v))$ is supermodular (submodular).

We are now ready for our central result:

**Theorem 3.3:** If $C_\theta(u,v)$ is a copula family that defines a positive PQD ordering, and the copula density $c_\theta(u,v)$ is TP2 for all values of $\theta$, then

$$\theta_1 < \theta_2 \Rightarrow -H(c_{\theta_1}) \leq -H(c_{\theta_2})$$

When $C_\theta(u,v)$ is RR2 the inequality is reversed.

**Proof:** Recall that $\mathbf{X} \sim c_{\theta_1}$ and $\mathbf{Y} \sim c_{\theta_2}$, and that $\theta_1 < \theta_2$. We will show that the following holds:

$$-H(\mathbf{X}) = \int c_{\theta_1}(u,v) log(c_{\theta_1}(u,v)) \mathrm{d}u \mathrm{d}v$$
$$\leq \int c_{\theta_2}(u,v) log(c_{\theta_1}(u,v)) \mathrm{d}u \mathrm{d}v$$
$$\leq \int c_{\theta_2}(u,v) log(c_{\theta_2}(u,v)) \mathrm{d}u \mathrm{d}v = -H(\mathbf{Y})$$

**First inequality.** Since $C_\theta(u,v)$ defines a PQD ordering, we have from Theorem 2.6 that $\mathbf{X} \leq_{sm} \mathbf{Y}$. Let $\Psi(u,v) = log(c_{\theta_1}(u,v))$. Since $c_{\theta_1}(u,v)$ is TP2, from Observation 3.2 we have that $\Psi(u,v)$ is super modular, that is $E(\Psi(\mathbf{X})) \leq E(\Psi(\mathbf{Y}))$. Thus:

$$\int c_{\theta_1}(u,v) \Psi(u,v) \mathrm{d}u \mathrm{d}v \leq \int c_{\theta_2}(u,v) \Psi(u,v) \mathrm{d}u \mathrm{d}v.$$

The first inequality follows by substitution of $\Psi$.

| Family | CDF | condition |
|---|---|---|
| Normal | $\Phi_\theta(\Phi^{-1}(u), \Phi^{-1}(v))$ | $0 \leq \theta \leq 1$ |
| FGM | $uv + \theta uv(1-u)(1-v)$ | $-1 \leq \theta \leq 1$ |
| Gumbel | $e^{-[(\hat{u})^\theta + (\hat{v})^\theta]^{1/\theta}}$ | $1 \leq \theta \leq \infty$, $\hat{u} = -log(u)$ |
| Frank | $-\frac{1}{\theta} \log\left(1 - \frac{\tau(u)\tau(v)}{\tau(1)}\right)$ | $0 \leq \theta \leq \infty$, $\tau(x) = 1 - e^{-\theta x}$ |
| Clayton | $\max(u^{-\theta} + v^{-\theta} - 1, 0)^{-\frac{1}{\theta}}$ | $\theta \in [-1, \infty], \neq 0$ |
| Joe | $1 - \left(\bar{u}^\delta + \bar{v}^\delta - \bar{u}^\delta \bar{v}^\delta\right)^{\frac{1}{\delta}}$ | $\delta \in [1, \infty)$, $\bar{u} = 1 - u$ |
| AMH* | $\frac{uv}{1 - \theta(1-u)(1-v)}$ | $-1 \leq \theta \leq 1$ |
| GB* | $uv e^{-\theta ln(u) ln(v)}$ | $0 \leq \theta \leq 1$ |

Table 1: TP2/RR2 Copula families. '*' marks families for which, to the best of our knowledge, this property was not previously known.

**Second inequality.** The difference between the two sides of the second inequality is

$$\int c_{\theta_2}(u,v) \left[ log(c_{\theta_1}(u,v)) - log(c_{\theta_2}(u,v)) \right] \mathrm{d}u \mathrm{d}v$$

The result follows by noting that this is simply the Kullback-Leibler divergence between the two densities $c_{\theta_2}$ and $c_{\theta_1}$, and the fact that this divergence is always non-negative [Cover and Thomas, 1991]. ∎

Note that the above theorem is stated for positively PQD ordered copula families. For negatively ordered families (e.g., Gumbel-Barnett) a reverse monotone relationship holds, as can be similarly proved.

The following is an immediate consequence of Theorem 3.3 and the known monotonicity of $\rho_s$ in the dependence parameter $\theta$ for PQD ordered families:

**Corollary 3.4 :** *If the copula density $c_\theta(u,v)$ is TP2/RR2 for all $\theta$, then the magnitude of Spearman's $\rho_s$ is monotonic in the copula entropy.*

Note that, phrased in terms of the *magnitude* of $\rho_s$, the result also holds for PQD families such as the Gaussian copula that are TP2 for one side of the parameter values ($0 \leq \theta \leq 1$) and RR2 otherwise ($-1 \leq \theta \leq 0$).

### 3.2 Examples of TP2/RR2 Copulas

As discussed, the density of many copulas is a TP2/RR2 function making our theoretical result widely applicable. Table 1 lists these copula families and provides their distribution function.

The Gaussian, Fairlie-Gumbel-Morgenstern (FGM), Frank, Gumbel, Clayton and Joe copulas are all known to have a TP2/RR2 density [Joe, 1997]. Thus, for all these families the negative entropy is monotonic in $\theta$, and consequently in the magnitude of Spearman's $\rho_s$.

Two additional popular copula families are confirmed to have a TP2/RR2 density:

**Lemma 3.5:** *The Ali-Mikhail-haq (AMH) copula has a TP2 density for nonnegative $\theta$ values and an RR2 density otherwise. The Gumbel-Barnett (GB) copula has an RR2 density.*

Proof of this result can be found in Appendix 7.

### 3.3 Other Copula Families

For completeness, we now discuss another sufficient condition for the monotonicity of the entropy in the dependence parameter and its relation to the TP2 condition. To the best of our knowledge, other than the work of Elidan [2012] that formulated the conjecture proved above, the only work that sheds theoretical light on the relationship between the copula dependence parameter $\theta$ and the entropy is that of Joe [1987]. The relevant details are summarized below.

If $f, g$ are two n-dimensional densities, then $f$ is said to be majorized by $g$, denoted by $f \preceq g$, iff $\int \Phi(f)dx \leq \int \Phi(g)dx$ for all convex functions $\Phi$. In particular, since $\Phi(x) = xlog(x)$ is convex, then letting $\mathbf{X}, \mathbf{Y}$ be two n-dimensional random vectors, such that $\mathbf{X} \sim f_1$, $\mathbf{Y} \sim f_2$, and $f_1$ is majorized by $f_2$, we have that

$$
\begin{aligned}
-H(\mathbf{X}) &\equiv \int f_1 log(f_1) d\mathbf{X} \\
&\leq \int f_2 log(f_2) d\mathbf{Y} \equiv -H(\mathbf{Y}).
\end{aligned}
$$

With some additional technical details (see [Joe, 1987]), it is possible to show that for all elliptical copula families the dependence parameter $\theta$ implies a majorization ordering, which in turn implies monotonicity of the entropy in the absolute value of Spearman's $\rho_s$.

Thus, the monotonicity of the entropy in the correlation parameter for a bivariate Gaussian copula can be proved via majorization or Theorem 3.3. However, majorization *does not* hold for the other families which have a TP2 density. Conversely, the t-copula elliptical family defines a majorization ordering but its density is neither TP2 nor RR2 for some degrees of freedom [Allan.R.Sampson, 1983].

Finally, the widely used Plackett family of copulas has a density that is neither a TP2 function, nor does it define a majorization ordering. However, as the simulations of Elidan [2012] suggest, the monotonicity of the entropy in the dependence parameter also holds for this copula family. Identification of the conditions necessary for the monotonicity relationship to hold remains a future challenge.

## 4 A Bayesian Approach for Learning Expressive Copula Trees

Base on the theoretical developments presented in the previous section, we now present a speedy model selection (SMS) approach for learning the structure of copula-based graphical models while allowing for different copula families within the same model. For clarity and simplicity of exposition, we focus on the case of copula trees where the relationship between the theory and practice is most direct. As we shall see in Section 5, even in this seemingly simple setting, our approach offers significant generalization benefits. We start with a brief review of a copula tree model and how Spearman's $\rho_s$ can be used to learn its structure for a single copula family.

### 4.1 Structure Learning using Spearman's $\rho_s$

We now we briefly review the idea put forth by Elidan [2012] for using Spearman's $\rho_s$ to learn the structure of a copula network, an idea whose theoretical substantiation has been greatly increased by the developments in the previous section.

In a tree structured copula model [Kirshner, 2007, Elidan, 2010], the joint density is represented as a product of bivariate copula densities corresponding to the edges of the tree $T$ and the univariate marginals:

$$f_\mathcal{X}(x_1, \ldots, x_n) = \prod_{(i,j) \in T} c_{ij}(F_i(x_i), F_j(x_k)) \prod_i f_i(x_i).$$

When learning the structure of a model, we seek a graph for which the (penalized) maximum likelihood function is highest. Since the likelihood function itself decomposes, the building block of learning is the evaluation of the merit of an edge $X \rightarrow Y$, independently of all other edges. In the case of the copula parameterization the relevant term is

$$Score(X, Y) \equiv \sum_{m=1}^M \log c_{\hat{\theta}}(F_X(x[m]), F_Y(y[m])),$$

where $\hat{\theta}$ are the estimated parameters, the sum is over training instances, and the marginal terms that do not depend on the graph's structure have been dropped.

Evaluation of $Score(X, Y)$ can be computationally difficult. However, all that we really need to identify the optimal tree is a *ranking* of the scores for all possible edges. If we assume that the data is generated from the copula, then as $M \rightarrow \infty$ we have that

$$Score(X, Y) \rightarrow -H(C_\theta(U, V)),$$

where $U, V$ are the ranks of $X, Y$, respectively.

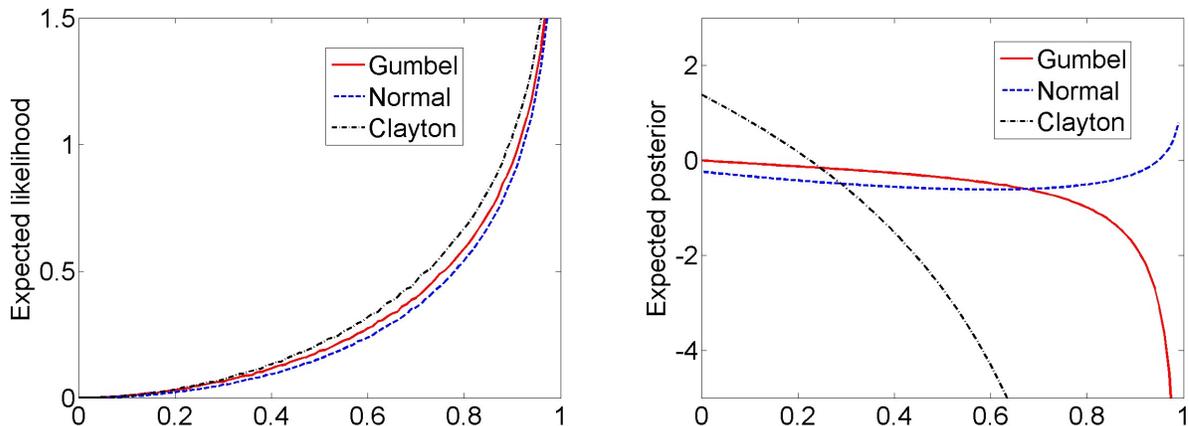

Figure 2: (left) expected log-likelihood vs. Spearman's $\rho_s$ for the normal, Gumbel and Clayton copula families. (right) expected posterior vs. Spearman's $\rho_s$.

Thus, having proved that $|\rho_s|$ is monotonic in $-H(C_\theta(U,V))$, we can simply use an easy to compute empirical estimate of $\rho_s$ to rank candidate edges, and find the optimal tree with respect to this measure using a simple maximum spanning tree algorithm.

### 4.2 Choosing From Multiple Copula Families

The above approach, suggested by Elidan [2012] allows for efficient structure learning of a copula based model that makes use of a *single* copula family. Obviously, not all pairs of variables share the same dependence characteristics and while the interaction between one pair of variable be be Gaussian, the interaction between another pair may be heavy-tailed and exhibit, for example, a behavior that is close to a Gumbel distribution. As an example, for the **Crime** census domain described in Section 5, the optimal tree includes around 41% edges parameterized by a Gaussian copula, 48% edges parameterized by a Gumbel copula, and 11% edges parameterized by a Clayton copula. Obviously, the need for a mix of copula families is real.

We now present an efficient approach for learning copula-based graphical models while allowing for a mix of copula families while retaining the lightning-speed efficiency of learning that is based on Spearman's $\rho_s$ empirical evaluation. Naively, since the expected log-likelihood is monotonic in Spearman's $\rho_s$, the following procedure may seem reasonable:

- Simulate the characteristic curve of expected log-likelihood for each copula family (note that this needs to be carried out only once).
- For each value of Spearman's $\rho_s$ choose the family that offers the highest expected log-likelihood

Unfortunately, such a procedure can fail since the theoretical result guarantees monotonicity *within* a copula family and not *between* copula families. In fact, as Figure 2(left) shows, the expected log-likelihood vs. Spearman's $\rho_s$ is highest for Clayton copula family through much of the range of $\rho_s$ values. Using the naive approach would lead us in this case to over-favor the Clayton copula.

An intuitive explanation to the above phenomenon is that while characteristic curves indeed capture the behavior *given* Spearman's $\rho_s$ for a particular family, they do not take into account the likelihood of seeing a particular value of $\rho_s$ within each family. In fact, we can expect the density of $\rho_s$ (which is always in the range $[-1, 1]$) to be quite different between the normal copula family whose dependence parameter is in the range $[-1, 1]$ and, for example, the Clayton copula family whose parameter has infinite support.

Given the above, we would like to somehow take into account a prior density over $\rho_s$ for each copula family. Using $\mathcal{C}$ to denote the set of copula families and $f_c(\rho_s)$ to denote the density of $\rho_s$ for a copula family $c \in \mathcal{C}$, we will then choose the copula family that maximize the expected posterior

$$argamax_{c \in \mathcal{C}} E\left(\log c(F(x), F(y); \rho_s)\right) + \ log f_c(\rho_s) \quad (3)$$

Importantly, this approach still relies on *precomputed* (posterior) characteristic curves and thus is as efficient as learning with a single copula family, regardless of the number of copula families considered.

The obvious question is how to choose the prior $f_c(\rho_s)$ for each copula family. In depth exploration of this question is left to future work and in here we use a straightforward approach which, as will be seen in Section 5, proves quite effective in practice. We choose a prior that, while taking into account the range of dependence parameter for each copula family, assigns higher density to the independence model where $\rho_s = 0$. Appealingly, this is both uninformative while at the same time ensuring that we do not encourage dependence that is due to finite data noise.

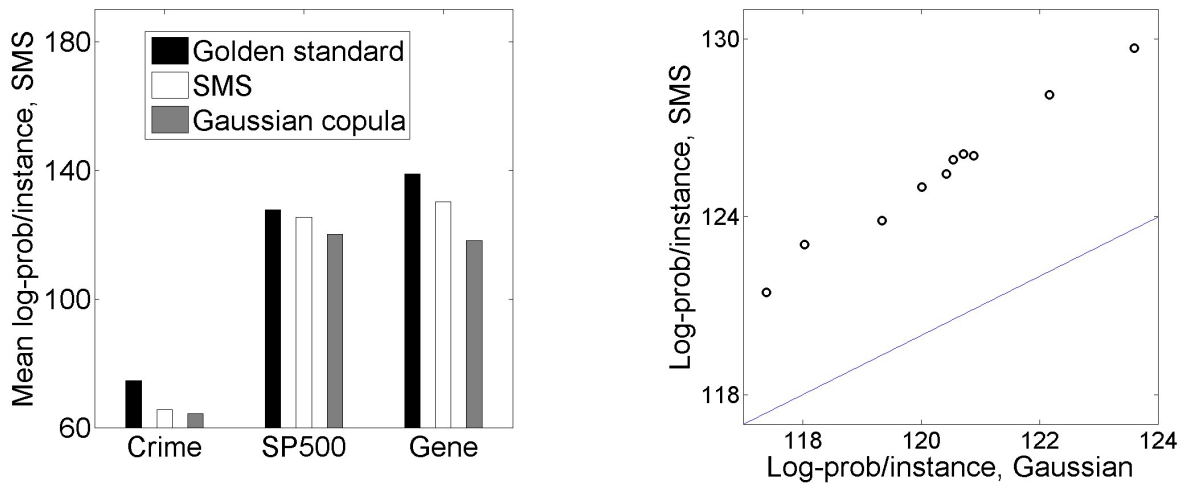

Figure 3: (left) Average log probability per instance over 10 folds. Compared are the Gaussian copula model, the model with a mix of copula families using our SMS method, and the golden standard of learning using exact maximum likelihood computations. Shown are results for the **Crime**, **SP500**, and **Gene** datasets. (right) comparison of our approach to the Gaussian copula baseline for all repetitions for the **SP500** domain.

Concretely, in this paper we consider the normal copula and the most popular Archimedean copula families, namely the Clayton, Frank, and Gumbel copula families. For the normal copula, the above translates into a standard truncated Laplace (also known as double exponential prior). For the Gumbel copula we use a shifted by unity exponential distribution (according to dependence parameter support) and for Clayton we take exponential distribution with parameter $\lambda = 4$. The resulting characteristic curves are shown in Figure 2(right) where it is clear that different copula families are preferred (highest) in different regions. In the next section we will show that using this curve to automatically choose the copula family based on empirical Spearman's $\rho_s$ evaluation results in competitive models that are learned very efficiently.

We note that the characteristic curve for the Frank copula family is missing from this graph because of its similarity to that of the normal copula family. The implication of this similarity is that our method cannot be used to separate these two copula families. This should not come as a great surprise since the symmetric Frank density, while not Gaussian, is much more similar to the normal distribution than, for example, the asymmetric heavy-tailed Gumbel one. While this may sound problematic, due to the similarity of densities, the practical implications of wrongly choosing between the Frank and Gaussian copulas are relatively small, as confirmed in preliminary experiments (not shown here for clarity of exposition).

## 5 Experimental Evaluation

In this section we demonstrate the practical benefit of our speedy model selection (SMS) method for learning expressive real-valued copula graphical models. As noted, we focus on tree structures where the learning task decomposes into the bivariate evaluation of the merit of each edge in the network individually. The significant generalization advantage of copula-based graphical models over the standard Gaussian BN has been demonstrated in the past [Kirshner, 2007, Elidan, 2010] (and confirmed for our datasets). For clarity, in here we focus on the *additional* advantage over models that involve only a single copula family.

For each edge we allow for a Gaussian, Clayton or Gumbel copula, with the prior for each family as defined in Section 4. As a baseline we consider learning only with a Gaussian copula, which is the strongest of all single family baselines. We also compare to the golden standard of learning using exact maximum likelihood computations. For the univariate marginals in all cases, we use standard kernel-based approach [Parzen, 1962] with the common Gaussian kernel (see, for example, [Bowman and Azzalini, 1997] for details).

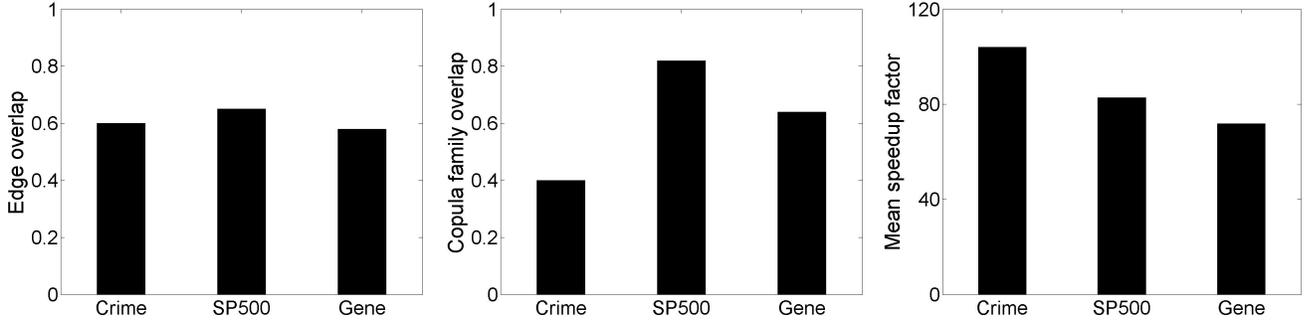

Figure 4: (left) Average fraction of edges that overlap between the model learned using our SMS method and the golden standard model learned using exact maximum likelihood computations. (middle) average fraction of edges, among those common to the two models, that agree on the copula family. (right) average speedup factor of our SMS method over learning with exact computations.

We consider three varied real-life datasets:

- **Crime** (UCI repository). 100 variables relating to crime ranging from household size to fraction of children born outside of a marriage, for 1994 communities across the U.S.
- **SP500**. End of day changes of the value of the 500 stocks (variables) comprising the Standard and Poor's index (S&P 500) over a period of close to 2000 trading days (samples).
- **Gene**. A compendium of gene expression experiments used in [Marion et al., 2004]. We chose genes that have at most one missing experiment. This resulted in 765 variables (genes) and 1088 samples.

Results for all three datasets are reported over 10 random equal splits into train and test samples.

We start by considering the average test log probability per instance, shown in Figure 3(left). The superiority of the mixed family model (white bars) over the Gaussian copula model (gray bar) is clear. Further, in the two bigger datasets, our highly efficient SMS approach improves substantially over the baseline both when taking into account the maximum likelihood golden standard and in absolute terms of bit per instance improvement. Appealingly, as the domain becomes more complex and the number of variable grows, so does our advantage. Figure 3(right) shows a typical more detailed comparison of performance for the **SP500** domain. As can be clearly seen, our superiority is consistently substantial over all random repetitions.

Next, we consider the qualitative ability of our SMS approach that is based on Spearman's $\rho_s$ evaluation to correctly identify both the structure and the best copula family for each edge. Figure 4 (left) shows for each of the datasets the average percentage of edges, over the 10 random runs, that are common to the model learned by our SMS method and the one learned using exact time-consuming computations. As can be clearly seen, the overlap between the trees learned is nontrivial considering the size of the domain the fact that our structure learning approach only relies on simple empirical Spearman's $\rho_s$ estimates. To evaluate the ability of our approach to also correctly choose the right parameterization for each edge, Figure 4 (middle) shows the average percentage of edges that, in addition to being in both the SMS model and that learned using exact computations, also agree on the copula family. Again, given the inherent difficulty of model selection, the similarity between the two models is appealing. The least favorable overlap for the crime domain also explains why our log-probability performance for the crime domain is the least impressive. Still, even in this case, performance is superior to the baseline Gaussian copula only model.

Finally, we consider the speedup factor of our SMS method when learning a mixed family copula model relative to learning using exact computations. Figure 4(right) shows the average speedup factor. For all three domains the speedup is quite impressive at around *two orders of magnitude*. Learning the **SP500** model, for example, takes only minutes on a single CPU making structure learning a significantly more accessible task than in the past. We note that the growth rate of both our SMS method and the exact one as a function of the number of variables is similar, and that the difference is in the dependence on the operations that have to be carried our for each training instance. The speedup reported confirms this since, for example, the **Gene** dataset has almost half the samples of the **Crime** dataset. Thus, while achieving impressive speedups even for the modest datasets considered here, our SMS method is particularly suited to handle a substantial number of training samples.

## 6 Summary and Future Work

In this paper we addressed the computationally demanding challenge of structure learning for real-valued domains in the context of expressive copula-based models. First, we significantly extended the result of Elidan [2012] and substantiated the conjecture of the monotonic relationship between the magnitude of Spearman's $\rho_s$ and the expected likelihood of an edge in the network. Second, we suggested a novel Bayesian approach for performing structure learning while also allowing for the selection of a different copula family for each edge, without incurring any computational cost. Third, we demonstrated the effectiveness of our SMS approach on varied real-life domains.

Importantly, the domains considered are quite sizable by structure learning standards and dramatically so for copula models. Further, to the best of our knowledge, ours is the first method for automated learning of multivariate copula-based models that allows for a mix of different copula families.

An obvious open theoretical question is the identification of *necessary* conditions for the monotonicity relationship between Spearman's $\rho_s$ and the copula entropy. Another important and practical avenue of research is the exploration of appropriate priors for the density of $\rho_s$ for different copula families. More generally, it would be useful to find other efficient proxies for speedy model selection, e.g., based on other dependence measures such as the Schwizer-Wolff sigma [Schweizer and Wolff, 1981].

## 7 Appendix

We now prove that the Ali-Mikhail-haq and Gumbel-Barnett copula densities are TP2/RR2, allowing us to apply Theorem 3.3 to these families. We start with a useful property of TP2 that will be used in our proof:

**Observation 7.1:** Let $f_1(x, y), f_2(x, y)$ be two real non-negative TP2 (RR2) functions. Then $\Psi(x, y) = f_1 f_2$ is TP2 (RR2).

**Proof:** From the TP2 property we have

$$\begin{aligned}
\Psi(x_1, y_1)\Psi(x_2, y_2) &= \\
&= f_1(x_1, y_1)f_1(x_2, y_2)f_2(x_1, y_1)f_2(x_2, y_2) \\
&\leq f_1(x_1 \vee x_2, y_1 \vee y_2)f_1(x1 \wedge x_2, y_1 \wedge y_2) \\
&\quad \times f_2(x_1 \vee x_2, y_1 \vee y_2)f_2(x1 \wedge x_2, y_1 \wedge y_2) \\
&\equiv \Psi(x1 \vee x_2, y_1 \vee y_2)\Psi(x_1 \wedge x_2, y_1 \wedge y_2),
\end{aligned}$$

where the last line follows from the previous by definition after rearranging of terms. The result follows from the definition of a TP2 density. ∎

**Lemma 7.2:** *The Ali-Mikhail-haq (AMH) density*

$$c(u,v) = \frac{1 + \theta[(1+u)(1+v) - 3] + \theta^2(\tilde{u})(\tilde{v})}{[1 - \theta(\tilde{u})(\tilde{v})]^{-3}},$$

*for $\theta \in [-1, 1]$ and where $\tilde{u} \equiv (1-u), \tilde{v} \equiv (1-v)$, is TP2 when $\theta \in [0, 1]$ and RR2 when $\theta \in [-1, 0)$.*

**Proof:** We will show that for $\theta \in [0, 1]$, the AMH density is a product of two non-negative TP2 functions, and the result will follow from Observation 7.1:

$$\begin{aligned}
f_1(u, v) &= 1 + \theta[(1+u)(1+v) - 3] + \theta^2 \tilde{u}\tilde{v}, \\
f_2(u, v) &= [1 - \theta\tilde{u}\tilde{v}]^{-3}.
\end{aligned}$$

The positivity of $f_1, f_2$, can be easily verified. Further, a positive function is TP2 iff it is supermodular on the log scale. Also, a function $f(u, v)$ is supermodular iff its second order derivatives are positive, that is $\partial u \partial v f(u, v) \geq 0, (u, v) \in \mathbf{R}^2$. We now have,

$$\frac{\partial^2 \log f_2(u, v)}{\partial u \partial v} = \frac{-3\theta^2 \tilde{u}\tilde{v} - \theta[1 - \theta\tilde{u}\tilde{v}]}{[1 - \theta\tilde{u}\tilde{v}]^2}.$$

This second order derivative is positive so that $f_2(u, v)$ is TP2. Using the same technique, we can show that $\log f_1(u, v)$ is supermoduler, hence $f_1(u, v)$ is also TP2. The proof for $\theta \in [-1, 0)$ is similar. ∎

**Lemma 7.3:** *The Gumbel-Barnett (GB) density by*

$$c(u, v) = \left(-\theta + [1 - \theta\grave{u}][1 - \theta\grave{v}]\right)e^{-\theta\grave{u}\grave{v}},$$

*for $\theta \in (0, 1]$, and where $\grave{u} = \theta log(1-u), \grave{v} = \theta log(1-v)$, is an RR2 function.*

**Proof:** We will show that the GB density is a product of two non-negative RR2 functions. Define

$$\begin{aligned}
f_1(u, v) &= -\theta + [1 - \theta\grave{u}][1 - \theta\grave{v}] \\
f_2(u, v) &= e^{-\theta\grave{u}\grave{v}}
\end{aligned}$$

$f_2$ is non-negative and since $c(u, v)$ is a density function, $f_1$ must also be non-negative. Also note that a function is RR2 only if is submodular on the log scale, and that a function is submodular if its second order derivative is non-positive. Starting with $f_1(u, v)$, we have

$$\frac{\partial^2 \log f_1(u, v)}{\partial u \partial v} = \frac{-\theta^3}{(1-u)(1-v)},$$

and this term is always non-positive for $\theta \in (0, 1]$. Thus, $f_1(u, v)$ is RR2. The proof that $f_2(u, v)$ is RR2 is similar. It follows that the Gumbel-Barnett density is an RR2 function. ∎

## Acknowledgements

This research is funded by ISF Centers of Excellence grant 1789/11, and by the Intel Collaborative Research Institute for Computational Intelligence (ICRI-CI).

# References

Allan.R.Sampson. Positive dependence properties of elliptically symmetric distributions. *Journal of Multivariate Analysis*, 13(2):375–381, 1983.

T. Bedford and R. Cooke. Vines - a new graphical model for dependent random variables. *Annals of Statistics*, 2002.

A. Bowman and A. Azzalini. *Applied Smoothing Techniques for Data Analysis*. Oxford University Press, 1997.

T. M. Cover and J. A. Thomas. *Elements of Information Theory*. John Wiley & Sons, New York, 1991.

G. Elidan. Lightning-speed structure learning of non-linear continuous networks. In *Proceedings of the AI and Statistics Conference (AISTATS)*, 2012.

Gal Elidan. Copula Bayesian networks. In *Advances in Neural Information Processing Systems (NIPS)*, 2010.

A.M. Hanea, D. Kurowicka, Roger M. Cooke, and D.A. Ababei. Mining and visualising ordinal data with non-parametric continuous bbns. *Comp Statistics and Data Analysis*, 54(3):668–687, 2010.

H. Joe. Majorization, randomness and dependence for mutivariate distributions. *The Annals of Probability*, 15(3):1217–1225, 1987.

H. Joe. Multivariate models and dependence concepts. *Monographs on Statistics and Applied Probability*, 73, 1997.

S. Kirshner. Learning with tree-averaged densities and distributions. In *Advances in Neural Information Processing Systems (NIPS)*, 2007.

D. Koller and N. Friedman. *Probabilistic Graphical Models: Principles and Techniques*. The MIT Press, 2009.

D. Kurowicka and R. Cooke. The vine copula method for representing high dimensional dependent distributions: Applications to cont. belief nets. In *Proc. of the Simulation Conf.*, 2002.

R.M. Marion, A. Regev, E. Segal, Y. Barash, D. Koller, N. Friedman, and E.K. O'Shea. Sfp1 is a stress- and nutrient-sensitive regulator of ribosomal protein gene expression. *Proc Natl Acad Sci U S A*, 101(40):14315–22, 2004.

R. Nelsen. *An Introduction to Copulas*. Springer, 2007.

E. Parzen. On estimation of a probability density function and mode. *Annals of Math. Statistics*, 33:1065–1076, 1962.

J. Pearl. *Probabilistic Reasoning in Intelligent Systems*. Morgan Kaufmann, 1988.

B. Schweizer and E. Wolff. On nonparametric measures of dependence for random variables. *The Annals of Statistics*, 9, 1981.

M. Shaked and J. Shanthikumar. *Stochastic Orders*. Springer, 2007.

A. Sklar. Fonctions de repartition a n dimensions et leurs marges. *Publications de l'Institut de Statistique de L'Universite de Paris*, 8:229–231, 1959.